\documentclass[conference,10pt]{ieeeconf}
\usepackage{cite}
\usepackage{graphicx}
\usepackage{float}
\usepackage{amsmath}
\usepackage{mathtools}
\graphicspath{ {images/} }
\usepackage{comment}
\usepackage[normalem]{ulem}
\useunder{\uline}{\ul}{}
\usepackage{subcaption}
\usepackage[tableposition=top]{caption}
\renewcommand\vec{\mathbf}

\usepackage{cite}
\usepackage{time}
\usepackage{comment}
\usepackage{amsmath}
\usepackage{mathabx}
\usepackage{multicol}
\usepackage[bookmarks=true]{hyperref}
\usepackage{graphicx}
\usepackage[utf8]{inputenc}
\usepackage{graphics} 
\usepackage{epsfig} 
\usepackage{mathptmx} 
\usepackage{bm}
\usepackage{caption}
\usepackage{copyrightbox}
\usepackage{subcaption}
\usepackage{ifthen}
\newboolean{include-notes}
\usepackage[linesnumbered,ruled,noend]{algorithm2e}
\usepackage[noend]{algpseudocode}
\newcommand{\mdnote}[1]{\ifthenelse{\boolean{include-notes}}%
 {\textcolor{DarkOrchid}{\textbf{MD: #1}}}{}}
\setboolean{include-notes}{true}

\makeatletter
\renewcommand{\CRB@setcopyrightfont}{%
\usefont{T1}{}{m}{n}\fontsize{9}{9}\selectfont
}

\makeatother
\renewcommand\vec{\mathbf}

\begin{document}

\title{Real-Time Online Re-Planning \\ for Grasping Under Clutter and Uncertainty}

\author{Wisdom C. Agboh \hspace{40mm} Mehmet R. Dogar \vspace{3mm} \\
School of Computing, University of Leeds \\
\{scwca, m.r.dogar\}@leeds.ac.uk
}

\maketitle

\begin{abstract} 
This work is published as a conference paper \cite{Agboh_Dogar_18} at IEEE Humanoids 2018. 

We consider the problem of grasping in clutter. While there have been motion planners
developed to address this problem in recent years, these planners are mostly
tailored for open-loop execution. Open-loop execution in this domain, however,
is likely to fail, since it is not possible to model the dynamics of the
multi-body multi-contact physical system with enough accuracy, neither is it
reasonable to expect robots to know the exact physical properties of objects,
such as frictional, inertial, and geometrical.  Therefore, we propose an online re-planning approach for grasping through clutter. The main challenge is the long planning times this domain requires, which makes fast re-planning and fluent execution difficult to realize. In order to address this, we propose an easily parallelizable stochastic trajectory optimization based algorithm that generates a sequence of optimal controls. We show that by running this optimizer only for a small number of iterations, it is possible to perform real time re-planning cycles to achieve reactive manipulation under clutter and uncertainty.
\end{abstract}


\section{Introduction}

In this paper, we consider the problem where a robot must reach through a cluttered
environment to grasp a target object. This problem is typically seen in warehouses where robots are required to retrieve items from shelves to fulfill a customer's order, or in our homes where a robot must reach into the fridge to pick up an object.  To do this, the robot needs to contact other objects in the environment and push them out of the way (Fig.~1a-d). An
object that is pushed by the robot may in turn push and dislocate other
objects, including the target object. Undesired events can happen during the
interaction, such as objects falling off the edge of the surface. 
The problem is the generation of robust and reactive robot actions that grasp the target object while preventing undesired events from taking place.

\begin{figure}
\centering
  \begin{subfigure}[b]{0.49\columnwidth}
  \centering
    \includegraphics[width=0.9\textwidth, height=0.9\textwidth]{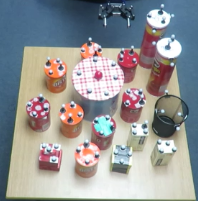}
    \begin{picture}(0,0)
    \put(10,48){\textbf{Target}}
    \put(40,48){\textbf{$\longrightarrow$}}
    \end{picture}
    \caption{Initial scene}
    \label{fig:fig1a}
  \end{subfigure}
  \hspace{-8mm}
  \begin{subfigure}[b]{0.49\columnwidth}
  \centering
    \includegraphics[width=0.9\textwidth, height=0.9\textwidth]{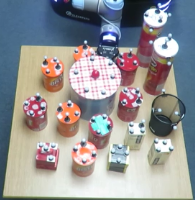}
    \caption{Contact triggers re-planning}
    \label{fig:2}
  \end{subfigure}
  \vspace{2mm}
  
  \begin{subfigure}[b]{0.49\columnwidth}
  \centering
    \includegraphics[width=0.9\textwidth, height=0.9\textwidth]{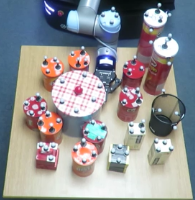}
    \caption{Re-planning}
    \label{fig:3}
  \end{subfigure}
  \hspace{-8mm}
  \begin{subfigure}[b]{0.49\columnwidth}
  \centering
    \includegraphics[width=0.9\textwidth, height=0.9\textwidth]{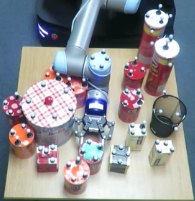}
    \caption{Target object grasped}
    \label{fig:4}
  \end{subfigure}
\caption{Snapshots from execution with online re-planning.}\label{fig:fig1}
\vspace{-6mm}
\end{figure}
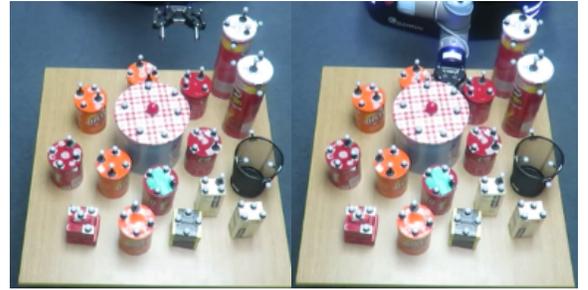
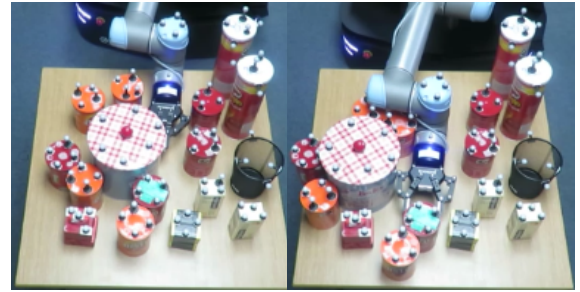

Existing work addresses this problem using motion planning followed by
\textit{open-loop} execution
\cite{kitaev_abbeel,king2015nonprehensile,dogar_clutter,randomized_clutter_uncertainty} i.e. the robot executes a sequence of actions one after the other without getting any feedback from the environment.
These approaches can be divided into two. The first approach uses motion
planning algorithms, e.g. kino-dynamic sampling-based algorithms
\cite{king2015nonprehensile} or trajectory optimization methods
\cite{kitaev_abbeel}, within a physics engine to generate the robot
trajectory. Trajectories that are produced this way, however, are likely
to fail in the face of uncertainty during real-world execution.  Consider
the scene in Fig.~\ref{fig:fig1a}, where the target object is near the center of the table. We can model this scene in a physics engine, plan a sequence of actions with a particular choice of physical parameters (e.g. friction
coefficients, object masses, object shapes) that take the robot to the
grasping goal state. However, if this plan is executed in an open-loop manner in
the real world, it can easily fail as the objects will not move exactly as predicted during planning. This is due to the uncertainty in the physics model of the physics engine and the assumed physical parameters of the objects.

The second open-loop approach addresses this problem by accounting for uncertainty during planning. This approach extends motion planners to generate actions that are robust to uncertainty
\cite{randomized_clutter_uncertainty,dogar_clutter,convergent_planning}. However, these planners are either restricted to a particular set of ``funneling'' actions, or generate highly conservative/pessimistic plans that are guaranteed to succeed under uncertainty. 

In this paper, we take a different reactive approach and investigate the potential
of closed-loop methods to address uncertainty during grasping in clutter.
One can use a planner to generate a plan to the goal state, execute a
portion of this plan, observe the environment, and then re-plan from
the resulting state to the goal, repeating this process until task
completion. This online re-planning or model predictive control (MPC)
approach has been implemented in many areas of robotics, including the
problem of pushing a single object \cite{pusher_slider,mppi_push}, but it
has not yet been explored for the problem of manipulation in clutter. 

The major challenge with online re-planning in this domain is long planning times. The average planning time reported in the
literature for the problem of grasping in clutter is in the order of
minutes
\cite{randomized_clutter_uncertainty,king2015nonprehensile,kitaev_abbeel,dogar_clutter}.
Then, under the online re-planning approach, the robot would need to execute a small action, update its world model with feedback and then will need to wait for possibly minutes before it receives
the next action from the planner. This long re-planning time makes it impractical for robots to use feedback from the environment in order to create new plans. Thus, this hampers real world applications and is highly
undesirable.

We propose an online re-planning approach to address this challenge.
First, we extend trajectory optimization methods that use
parallel trajectory rollouts \cite{mppi,kalakrishnan2011stomp} in search
of a lower-cost trajectory. By performing each roll-out on a different
core, we are able to reduce the time each iteration of our planner takes
to be equivalent to a single roll-out. Second, we track the deviation of
the actual state from the predicted state, and perform re-planning only if
the state deviation exceeds a threshold. This prevents us from planning at
every time step and allows us to have an automatic system that can be adjusted between open-loop execution and standard model predictive control. Finally, we formulate the problem as optimizing a cost function where reaching the goal is not a hard constraint, and therefore even if a quick re-planning cycle does not produce a trajectory that reaches the goal (i.e. grasps the target object) within the given
time limit, we can still use it if it is a lower-cost trajectory. 

Our specific contributions include an on-line re-planning (OR) algorithm to address uncertainty during grasping in clutter. We show that using our approach, one can achieve real time re-planning cycles with a robot in difficult and cluttered real environments. Real robot experimental results can be seen at \url{https://youtu.be/RcWHXL2vJPc}. Moreover, we compare OR to open-loop execution, particularly to \textit{naive re-planning} (NR), which plans a trajectory, executes it open-loop until the end, checks if the goal is achieved, and repeats this process if not. We show that OR is more successful in grasping the target object in a time limit, produces lower cost execution trajectories, and is faster. 

\section{Problem Definition}

As shown in Fig.~\ref{fig:fig1}, we consider the problem where a
robot must plan a trajectory from a given initial pose to a final pre-grasping
pose to retrieve an item from a cluttered environment. 
We consider a  planar robot consisting of an arm and a gripper as shown in the figure. The robot's state is defined by a vector of joint values $\vec{q}^{R} = \{\theta_x, \theta_y, \theta_{rotation}, \theta_{gripper}\}$,  where the $\theta$ values represent the x-axis prismatic joint, the y-axis prismatic joint, the rotational joint and the gripper's opening joint values, respectively. The scene includes $D+1$ movable dynamic objects. $\vec{q}^i$ refers to the six-dimensional pose (three translations and three rotations) of each object, for $i={1,\ldots,D}$.
$\vec{q}^{Target}$ refers to the pose of the target object, i.e. the object to be grasped. 
We assume a flat surface with edges, such as the table in Fig.~\ref{fig:fig1}, 
and dropping any object off the edges is undesired.

We use $\vec{x}_{t}$ to represent the complete state of our system at time $t$, which
includes the state of the robot and all objects; ${\vec{x}_{t} =\{\vec{q}^{R},\vec{q}^1,\ldots,\vec{q}^D,\vec{q}^{Target}\}}$. We consider a control input $\vec{u}_t$ applied at time $t$ for a fixed
duration $\Delta_{t}$. The controls in our case are velocities applied to the robot's degrees of freedom; ${\vec{u}_{t}=\{\dot{\theta}_x,\dot{\theta}_y, \dot{\theta}_{rotation}, \dot{\theta}_{gripper}\}}$.
Then, the discrete time dynamics  of the system is defined as: 
\begin{equation}
 \vec{x}_{t+1} = f(\vec{x}_{t}, \vec{u}_t)
\end{equation}
where $f$ is the state transition function.

We assume an initial state of the system, $\vec{x}_0$, and we define our goal as
generating a sequence of control inputs, such that the gripper grasps the
target object as quickly as possible, without dropping objects off the table.
We use the notation $\vec{u}_{0:n-1}$ to represent a sequence of control signals
through $n$ time steps, each applied for a fixed duration.  Similarly, we use $\vec{x}_{0:n}$ to represent a sequence
of states.

We use a physics engine \cite{mujoco} simulating rigid-body dynamics to model
$f$. Nevertheless, any physics engine is an inaccurate model of the real-world physics and uncertainties over the system dynamics are inevitable. Indeed even if we assumed perfect modeling, it is difficult for a robot to know the exact 
geometric, frictional, and inertial properties of objects in an environment.
In addition, object tracking systems come with inaccuracies in the estimation
of object poses in an environment.
Therefore, our objective in this work is to find a sequence of controls that
would move the system to a goal state even under an inaccurate model of the
system and its dynamics.
\section{Proposed Approach}

To address the inaccuracies mentioned above, we propose to use an online
re-planning approach, where the robot makes a plan, executes a portion of it,
observes the resulting state, and re-plans. 

Below, in Sec.~\ref{sec:pbsto}, we first present the planner that we use to
generate a sequence of controls to the goal from a given state. In
Sec.~\ref{sec:or}, we show how we use this planner within an online re-planning
framework. In Sec.~\ref{sec:nr}, we present the baseline 
approach we compare against in this paper.

\subsection{Physics-based trajectory optimization}\label{sec:pbsto}

Recent stochastic trajectory optimization methods such as
STOMP \cite{kalakrishnan2011stomp} and model predictive control methods such
as MPPI \cite{mppi} show impressive speed by using parallel rollouts.
Moreover, since these are optimization-based methods, even when they are used with 
a small time limit, they can still output an improved lower-cost trajectory,
even if the trajectory is not necessarily reaching a goal state. 
In contrast, sampling-based planners such as RRTs and PRMs
\cite{king2015nonprehensile,randomized_clutter_uncertainty} typically do not return a useful solution unless they are run until a path to the
goal is found, which can take minutes.
To the best of our knowledge, such parallelizable stochastic trajectory
optimization methods have not yet been used to solve grasping in clutter problems.
However, the properties we
mention above make parallelizable stochastic trajectory optimization methods
a promising approach for online re-planning to address problems in this domain.

\begin{figure}
\centering
\begin{minipage}{.5\columnwidth}
  \centering
  \vspace{3mm}
  \includegraphics[width=.4\linewidth, height=0.6\linewidth]{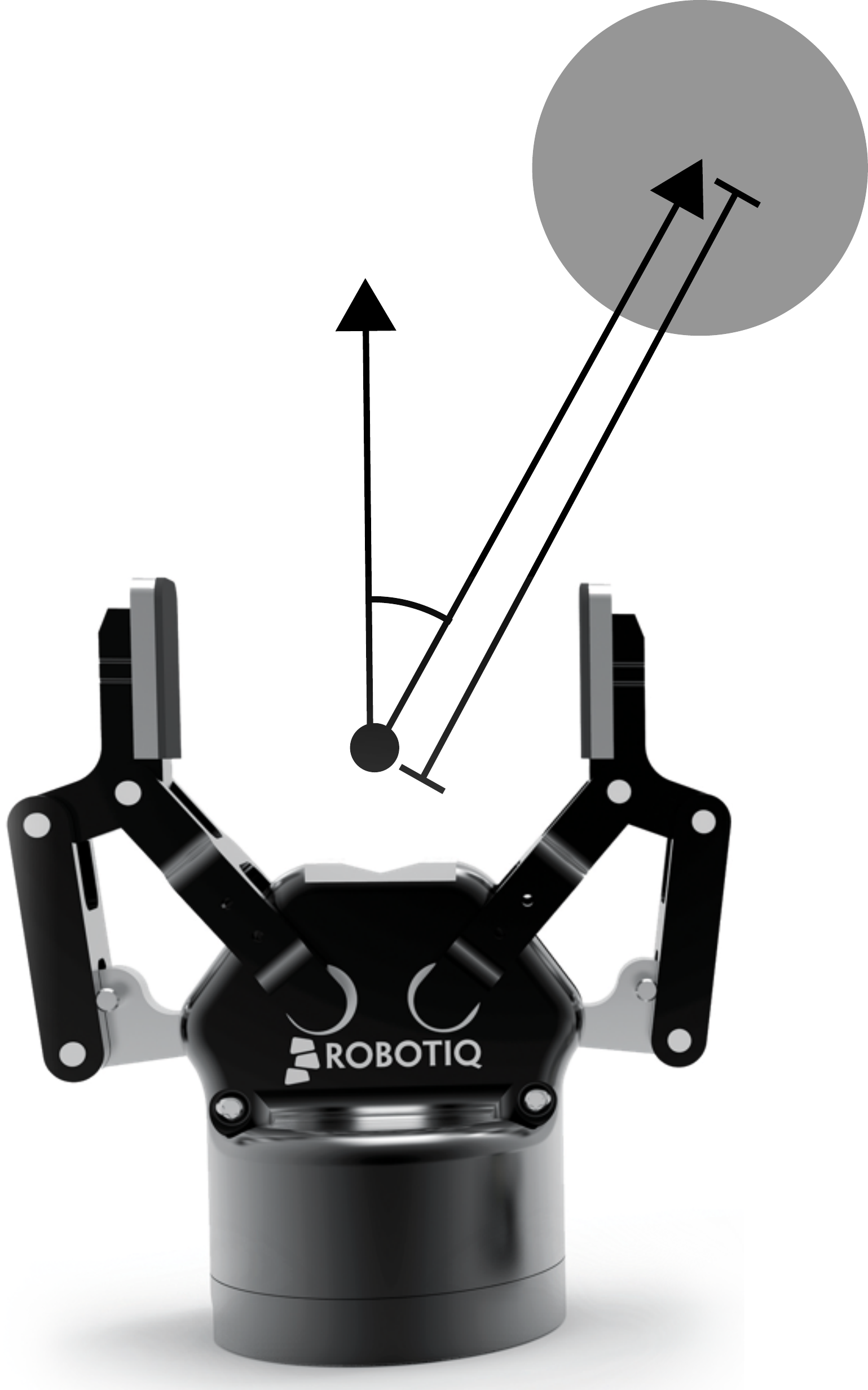}
     \begin{picture}(0,0)
      \put(0.0,67.0){Target}
      \put(-30.0,52.0){\small{$\phi_T$}}
      \put(-16.5,47.2){$d_T$}
    \end{picture}
  \captionof{figure}{Goal cost terms}
  \label{fig:goalcost}
\end{minipage}%
\begin{minipage}{.5\columnwidth}
  \centering
    \vspace{3mm}
  \includegraphics[width=.63\linewidth, height=0.6\linewidth]{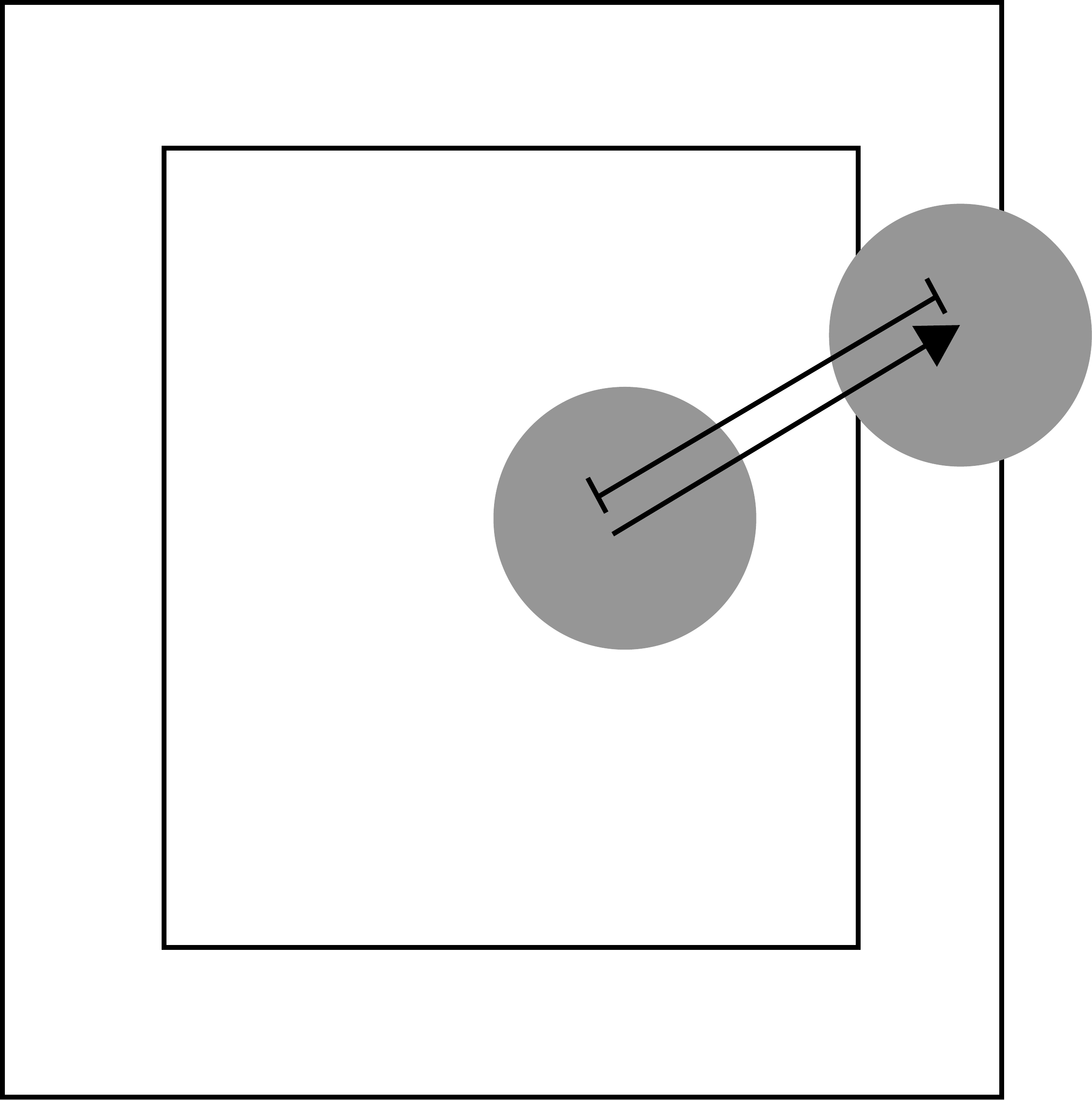}
     \begin{picture}(0,0)
      \put(-65.0,14.0){Safe zone}
      \put(-55.0,1.0){Table}
      \put(-33.0, 54.0){$d_E^i$}
      \put(-55.0,30.0){$\vec{x}_{t}^{i}$}
      \put(-7.0,37.0){$\vec{x}_{t+1}^{i}$}
    \end{picture}
  \captionof{figure}{Edge cost terms}
  \label{fig:edgecost}
\end{minipage}
\vspace{-6mm}
\end{figure}

\noindent We formulate the following problem: 
\begin{align}
    \min \limits_{\vec{u}_{0:n-1}} [ w_g \cdot c_g(\vec{x}_{n}) + \sum_{t=0}^{n-1} \sum_{r}^{} (w_{a} \cdot c_{a} + w_{d} \cdot c_{d} +  w_{e} \cdot c_{e}  )]
\end{align}
\vspace{-0.5cm}
\begin{align*}
    s.t. \hspace{3mm}  \vec{x}_{t+1} = f(\vec{x}_{t}, \vec{u}_{t}), 
\end{align*}
\vspace{-0.5cm}
\begin{align*}
    \hspace{7mm} \vec{x}_{0} \hspace{1.5mm} is \hspace{1.5mm} fixed, \hspace{3mm} \vec{u}_{t}=0 \hspace{1.5mm} for \hspace{1.5mm} t<0.
\end{align*}
where we search for an optimal sequence of controls $\vec{u}_{0:n-1}$ that
minimizes the weighted combination of costs.  We use four cost terms; $c_{g}, c_{d}, c_{e}, c_{a}$ and corresponding weights $w_{g}, w_{d}, w_{e}, w_{a}$.

\begin{itemize}
    \item $c_g(x_n) = d_{T}^2+w_{\phi}\cdot \phi_{T}^2$. This is the
        terminal \textit{goal cost} term,
    quantifying how far the robot hand is from grasping the target object at the final state.
    We illustrate how the distance $d_{T}$ and the angle
    $\phi_{T}$ are computed in Fig.~\ref{fig:goalcost}. 
    We first draw a vector from a fixed point in the gripper to the target object.
    $d_T$ is the length of this vector, i.e. the distance between the fixed point in the gripper and the target object.
    $\phi_{T}$ is the angle between the forward direction of the gripper and the vector.
    We use $w_{\phi}$ to weight angles relative to distances. 
    \item $c_d(\vec{u}_{t-1:t} , \vec{x}_{t:t+1}) = \sum_i^D (\vec{x}_{t+1}^i - \vec{x}_{t}^i)^2$. This is
    the \textit{disturbance cost} term, quantifying how much each object moved
    between two time-steps. This term encourages the robot to minimize the
    change in the configuration of the
    rest of the scene.
\item $c_e(\vec{u}_{t-1:t} , \vec{x}_{t:t+1}) = \sum_{i} e^{\{k\cdot d_E^i\}}$ for all $i$ out
    of the safe zone. This is the \textit{edge cost} term, penalizing 
    those objects that get too close to the boundary of the table or that get
    out of the boundary. As we illustrate in Fig.~\ref{fig:edgecost}, we define
    a safe zone that is smaller than the boundary of the table. If at time
    $t+1$ an object $i$ is out of this safe zone, we compute the distance it
    is pushed between $t$ and $t+1$, which we define as $d_E^i$. $k$ is a constant term.
    We do not add any edge costs for objects that are in the safe zone.
    \item $c_a(\vec{u}_{t-1:t} , \vec{x}_{t:t+1}) = (\vec{u}_t - \vec{u}_{t-1})^2$. This is the
    \textit{acceleration cost} term, with which we penalize large changes
    in robot velocities between two time steps.
\end{itemize}

Note that, instead of imposing the terminal grasping state as a hard
constraint, we declare it as a cost term,
$c_g$. We are able to accept trajectories that do not reach the goal completely, because we use this planner in a re-planning framework, i.e. we can rely on future re-planning cycles to take us to the goal. 

We solve this problem using Alg.~\ref{alg:pbsto}, which adapts the STOMP
algorithm \cite{kalakrishnan2011stomp} for physics-based grasping through
clutter.

\begin{algorithm}\label{alg:pbsto}
    \SetKwInOut{Input}{Input}
    \SetKwInOut{Output}{Output}
    \SetKwInOut{Parameters}{Parameters}
    \SetKwInOut{Subroutines}{Subroutines}
    \Input{
     $\vec{x}_{0}$: Initial state \\ 
     $\vec{u}_{0:n-1}$: Initial control sequence \\
     $I_{max}$: Maximum number of iterations }
    \Output{$\vec{u}_{0:n-1}$: Control sequence\\
            $\vec{x}_{0:n}$: Predicted states}
    \Parameters{
     $K$: Number of noisy trajectory rollouts \\
     $\nu$: Sampling variance \\
     $C_{thresh}$: Cost threshold implying success \\
     $n_{min}$: Minimum number of time steps}
     \Subroutines{
     $Cost$: Computes total cost, i.e. the minimized value in Eq. (2).}
     $\vec{x}_{0:n} \gets$ Roll out $\vec{u}_{0:n-1}$ over $\vec{x}_0$ to get initial state sequence \\
     \While{$I_{max}$ not reached and $Cost(\vec{u}_{0:n-1},\vec{x}_{0:n}) > C_{thresh}$}{
     \For{$k \gets 0 $ \KwTo $K-1$ }
     {
     $\vec{x}^k_{0} \gets \vec{x}_{0}$ \\
     $\vec{u}^k_{0,n-1} \gets N(\vec{u}_{0:n-1},\nu)$\\
     \For{$t \gets 0$ \KwTo $n-1$}
     {
     $\vec{x}^k_{t+1} \gets f(\vec{x}^k_{t}, \vec{u}^{k}_{t})$ \\
     \If{$Cost(\vec{u}^k_{0:t},\vec{x}^k_{0:t+1}) \leq C_{tresh}$ and  $t \geq n_{min} $ }
     {\Return ($\vec{u}^k_{0:t}$,$\vec{x}^k_{0:t+1}$)}
     }
     }
     $k^* \gets arg \min \limits_{k} (Cost(\vec{u}^{k}_{0:n-1},\vec{x}^k_{0:n}))$ \\
   
     \If {$Cost(\vec{u}^{k^*}_{0:n-1},\vec{x}^{k^*}_{0:n}) < Cost(\vec{u}_{0:n-1},\vec{x}_{0:n})$ } 
     {
         $\vec{u}_{0:n-1} \gets \vec{u}^{k^*}_{0:n-1}$ \\
         $\vec{x}_{0:n} \gets \vec{x}^{k^*}_{0:n}$ 
     }
     
 }\Return $(\vec{u}_{0:n-1},\vec{x}_{0:n})$
 
    \caption{Physics-Based Stoch. Traj. Optim. (PBSTO)}\label{alg:pbsto}
\end{algorithm}

We start with an initial candidate control sequence $\vec{u}_{0:n-1}$. During each iteration between lines 2-13, we try to improve this control sequence, until the cost is lower than a threshold, or until a maximum number of iterations is reached (Line 2). During each iteration, we create $K$ new control sequences, roll out these
controls in parallel using our model of the system, and compute the cost for each (Lines 3-9). Each new control sequence $\vec{u}^k_{0:n-1}$ is created by adding stochastic noise to the candidate control sequence $\vec{u}_{0:n-1}$ (Line 5).  The control sequence with the minimum cost is then identified and set as the new candidate control sequence.

Most robot motion planners that use trajectory optimization formulate the
problem as a fixed horizon problem, i.e. with a predetermined number of
time-steps/way-points. In the problem of grasping in clutter however, the length
of the required trajectory can change significantly: For example, the target
object may be pushed and moved away from its initial position, and this may
require a much longer trajectory than a case where the target is grasped at its
original position. Therefore, we initialize the planner with a long enough
control sequence, but also allow it to short-cut trajectories if the cost
indicates success earlier (Lines 8-9). Moreover, physics-based trajectory roll
outs are time consuming, hence truncating the roll out when success has been
achieved leads to lower planning times.

\subsection{Online Re-planning}\label{sec:or}
If allowed to run for many iterations, i.e. with a large $I_{max}$,
Alg.~\ref{alg:pbsto} can generate successful plans for the problem of grasping under clutter, as we show in our results in Sec.~\ref{sec:experiments_and_results}. 
However, when executed open-loop, these plans are likely to fail due to the uncertainties in the system dynamics, inaccuracies in the physical properties of the objects, and the state observations. To address this, we use Alg.~\ref{alg:pbsto} within an online re-planning (OR)
algorithm, which we present in Alg.~\ref{alg:or}.  
\vspace{-2mm}
\begin{algorithm}
      \SetKwInOut{Input}{Input}
    \SetKwInOut{Output}{Output}
    \SetKwInOut{Parameters}{Params}
    \Input{
        $\vec{u}_{0:n-1}$: Initial controls, e.g. straight line motion
    }
    \Parameters{
    $SD_{thresh}$: State deviation threshold \\
    $n_{min}$: Minimum number of controls to optimize\\
    $ManyIter$: Large number of iterations, e.g. 50 \\
    $FewIter$: Small number of iterations, e.g. 1 \\
    }
    $\vec{x}_{current} \gets $ Observe current state \\
    $(\vec{u}_{0:n-1},\vec{x}_{0:n}) \gets $ PBSTO($\vec{x}_{current}$,$\vec{u}_{0:n-1}$,$ManyIter$) \\
  
    \While{target object not grasped}{
		Execute $\vec{u}_{0}$ \\
        Remove $\vec{u}_0$ from sequence, i.e. $\vec{u}_{0:n-2} \gets \vec{u}_{1:n-1}$ \\
        Remove $\vec{x}_0$ from sequence, i.e. $\vec{x}_{0:n-1} \gets \vec{x}_{1:n}$ \\
        $\vec{x}_{current} \gets $ Observe current state \\

	    \eIf{ target object not predicted to be grasped at $\vec{x}_n$ \\
            \textbf{or} large state deviation, i.e. $||\vec{x}_{0}-\vec{x}_{current}|| > SD_{thresh}$ \\
            \textbf{or} too few controls left, i.e. $n-1 < n_{min}$}
    	{	
            $\vec{u}_{0:n-1} \gets \vec{u}_{0:n-2}$ $+$ single straight step to target \\
            $(\vec{u}_{0:n-1},\vec{x}_{0:n}) \gets $ PBSTO($\vec{x}_{current}$,$\vec{u}_{0:n-1}$,$FewIter$) \\
        }{
            $n \gets n-1$ \Comment{Decrement length of controls}
        }
    }
  \caption{Online Re-planning (OR)}\label{alg:or}
\end{algorithm}
\vspace{-2mm}

On line 2, we generate a locally optimal open-loop trajectory by calling the PBSTO planner with a large number of iterations. Then we start executing this trajectory. After
execution of every control action (line 4), we observe the current state (line
7), and then  re-plan from this current state (line 12).  However, when we
re-plan, we call the planner with only a few iterations, to receive fast, close
to real-time, updates to the plan. We warm-start the trajectory optimizer by
providing the previous plan. Furthermore, we re-plan only if it is necessary.
To do this, we check if the final predicted state of the current plan grasps
the target object (line 8), and we check if there are too few controls left in
the plan (line 10). More importantly, if the real observed state is evolving
according to the planner's predictions, and the other previously mentioned conditions are still satisfied, we do not re-plan. We check this on line
9, where we compute the deviation between the observed state and the first state
of the planned trajectory, and verify if this deviation is less than a threshold.
This threshold can be used to adjust how reactive the system is to unexpected events.

\subsection{Naive Re-planning}\label{sec:nr}
Open-loop execution during grasping in clutter can be unsuccessful due to uncertainty. In this paper we propose to address this problem through online feedback control. However, a naive approach to fixing this problem can be re-planning if success is not achieved after the complete open-loop execution of a plan.  We present this Naive Re-planning (NR) approach in Alg.~\ref{alg:nr}, and use it as a baseline in our experiments.
\vspace{-2mm}
\begin{algorithm}
    \SetKwInOut{Input}{Input}
    \SetKwInOut{Output}{Output}
    \SetKwInOut{Parameters}{Params}
    \Parameters{
    $ManyIter$: Large number of iterations, e.g. 50 \\
    }
    \While{target object not grasped}{
        $\vec{x}_{current} \gets $ Observe current state \\
        $\vec{u}_{0:n-1}\gets$ initial controls, e.g. straight to target object\\
        $(\vec{u}_{0:n-1},\vec{x}_{0:n}) \gets $ PBSTO($\vec{x}_{current}$,$\vec{u}_{0:n-1}$,$ManyIter$) \\
		Execute $\vec{u}_{0:n-1}$ \\
    }
 \caption{Naive Re-planning (NR)}\label{alg:nr}
\end{algorithm}
\vspace{-6mm}
\section{Experiments and Results}
\label{sec:experiments_and_results}

Through our experiments, we compare the online re-planning (OR) approach with
the naive re-planning (NR) approach.  We hypothesize that OR is more successful
in grasping the target object, that OR results in an execution cost that is
lower-cost, and that OR is also faster. We investigate whether using the
physics-based stochastic trajectory optimization (PBSTO) method, we can
reactively re-plan close to real-time, or at least fast enough to avoid
noticeable delays during execution.

We implemented our algorithms using the Mujoco \cite{mujoco} physics
engine. We perform experiments both in simulation and on a real robot.
As shown in Fig.~\ref{fig:fig1}, we assume a world consisting of objects on a table, and a planar robot with a two finger gripper. We make a distinction between two different type of worlds we deal with. 

\noindent\textbf{Planning world}: The planning world is a simulation
environment where the robot generates its plans/controls.  

\noindent\textbf{Execution world}: The execution world is the environment where
the robot executes actions and observes the resulting actual state. The
execution world is simulated for the simulation experiments and it is the
physical world for real robot experiments. 

Whether in simulation or on the real robot, we assume a mismatch between the
physics of the execution world and the planning world, the physical object
properties of the two worlds, and the state of the two worlds. We use the term
\textit{uncertainty level} to refer to the degree of this mismatch.  For
example, \textit{no uncertainty} implies a perfect match between the Planning
World and the Execution World, which is only possible in simulation experiments.
\textit{Low uncertainty} implies a low level of mismatch, and so on.

\begin{figure*}[htb!]
  \centering
  \begin{subfigure}[b]{0.245\textwidth}
      \copyrightbox[l]{\includegraphics[height=1.20in,width=1.20in]{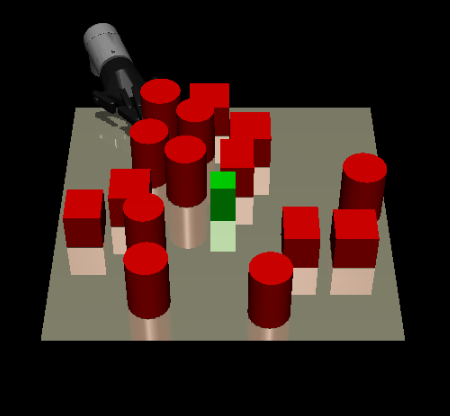}}{Planned motions}
    \end{subfigure}
  \hspace{-15.1mm}
  \begin{subfigure}[b]{0.245\textwidth}
    \copyrightbox[l]{\includegraphics[height=1.20in,width=1.20in]{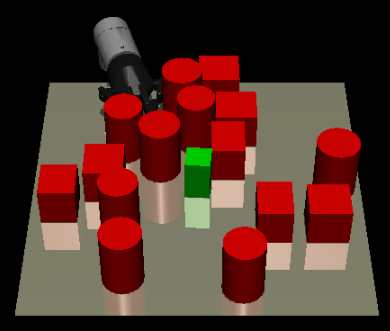}}{}
  \end{subfigure}
  \hspace{-16.5mm}
     \begin{subfigure}[b]{0.245\textwidth}
    \copyrightbox[l]{\includegraphics[height=1.20in,width=1.20in]{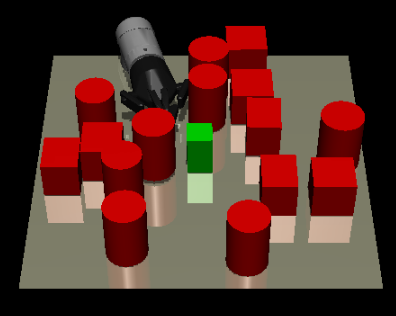}}{}
  \end{subfigure}
  \hspace{-16.5mm}
  \begin{subfigure}[b]{0.245\textwidth}
    \copyrightbox[l]{\includegraphics[height=1.20in,width=1.20in]{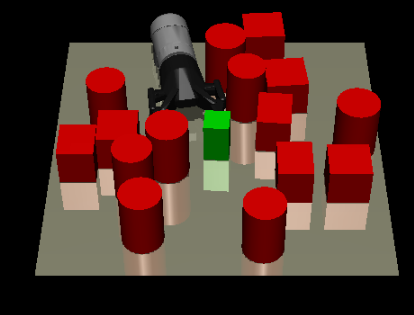}}{}
  \end{subfigure} 
  \vspace{-5mm}
  \begin{subfigure}[b]{0.245\textwidth}
  \vspace{-5mm}
      \copyrightbox[l]{\includegraphics[height=1.20in,width=1.20in]{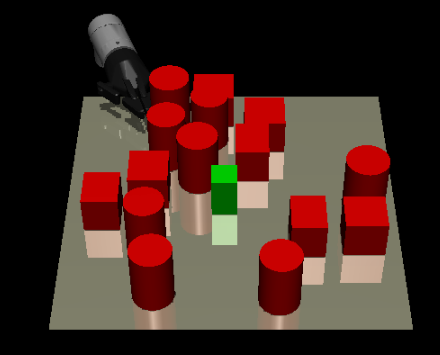}}{Execution with NR}
  \end{subfigure}
  \hspace{-15.1mm}
  \begin{subfigure}[b]{0.245\textwidth}
    \copyrightbox[l]{\includegraphics[height=1.20in,width=1.20in]{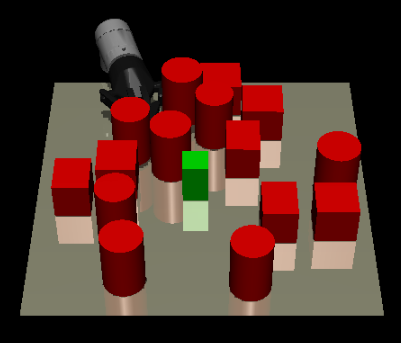}}{}
  \end{subfigure}
    \hspace{-16.5mm}
  \begin{subfigure}[b]{0.245\textwidth}
    \copyrightbox[l]{\includegraphics[height=1.20in,width=1.20in]{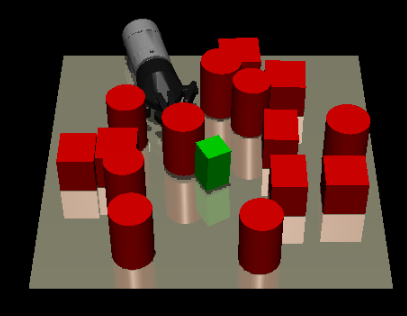}}{}
  \end{subfigure}
  \hspace{-16.5mm}
  \begin{subfigure}[b]{0.245\textwidth}
    \copyrightbox[l]{\includegraphics[height=1.20in,width=1.20in]{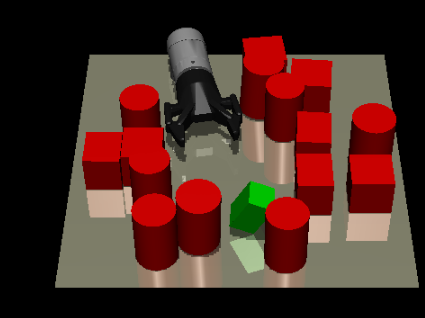}}{}
  \end{subfigure} 
%
%
  \begin{subfigure}[b]{0.245\textwidth}
      \copyrightbox[l]{\includegraphics[height=1.20in,width=1.20in]{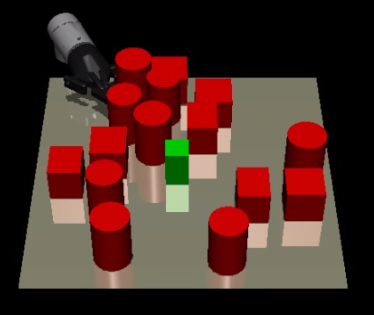}}{Execution with OR}
  \end{subfigure}
  \hspace{-15.1mm}
  \begin{subfigure}[b]{0.245\textwidth}
    \copyrightbox[l]{\includegraphics[height=1.20in,width=1.20in]{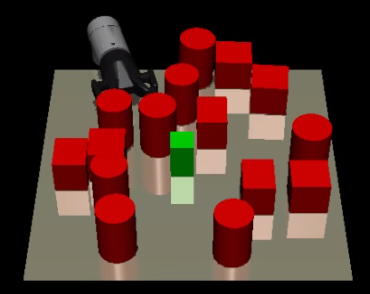}}{}
  \end{subfigure}
    \hspace{-16.5mm}
  \begin{subfigure}[b]{0.245\textwidth}
    \copyrightbox[l]{\includegraphics[height=1.20in,width=1.20in]{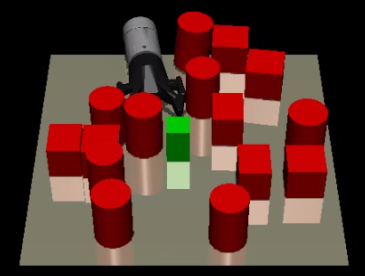}}{}
  \end{subfigure}
  \hspace{-16.5mm}
  \begin{subfigure}[b]{0.245\textwidth}
    \copyrightbox[l]{\includegraphics[height=1.20in,width=1.20in]{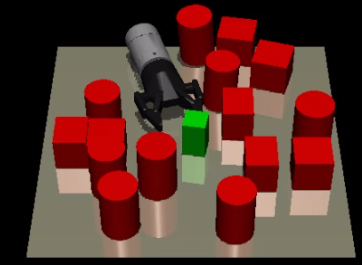}}{}
  \end{subfigure} 

\caption{Top row: The planned control sequence and state
evolution. Middle row: Open-loop execution of the planned control sequence
fails under medium uncertainty. Bottom row: Online re-planning algorithm OR
succeeds under medium uncertainty. }\label{fig:examplesims}
\end{figure*}

\subsection{Simulation experiments}
We perform experiments in simulation to evaluate the performance of our
planners in scenes with varying degrees of clutter and uncertainty. 
We begin by creating \textit{execution worlds}. Here,
the \textit{execution world} is created in Mujoco and it consists of 15 objects
(boxes and cylinders), a $0.6 m \times 0.6 m$ table and our planar
robot as shown in Fig.~\ref{fig:examplesims}. 

For each execution world: 
\begin{itemize}
\item We randomly select a shape (box or cylinder) for each of the 15 objects.
\item For each object, we randomly select\footnote{
The uniform range used for each parameter is given here. 
Box x-y extents: $\left[0.03m,0.05m\right]$; 
box height: $\left[0.036m,0.04m\right]$;
cylinder radius:$\left[0.035m,0.04m\right]$;
cylinder height:$\left[0.04m,0.055m\right]$;
mass:$\left[0.2kg,0.8kg\right]$;
coef. fric.:$\left[0.2,0.6\right]$.
} 
shape dimensions (extents for the
boxes, radius and height for the cylinder), mass, and coefficient of friction.
\item We select a pose for the target object from a Gaussian with a mean at the center of the table and a variance of $0.01 m$.
\item For the other 15 objects, we randomly select non-colliding object poses on the table.
\end{itemize}

We generate 100 such execution worlds. To generate a planning world from an execution world, we add Gaussian noise onto the physical parameters of the execution world\footnote{
The variance of the Gaussian noise for each parameter under low-uncertainty are given here. These values are multiplied by $2$ for medium, and $3$ for high uncertainty.
Object pose translation: $0.005$;
Object pose rotation around vertical axis: $0.005$;
Box x-y extents, cylinder radius, and height: $0.005$; 
mass:$0.01$;
coef. fric.:$0.005$.
}. For each execution world, we create four such planning worlds with increasing amounts of noise, corresponding to the four uncertainty levels: no uncertainty, low, medium, and high uncertainty. Given a pair of Planning world and Execution world, we then run and execute one
of our planners. Moreover, we simulate physics stochasticity in the execution world by adding 
Gaussian noise\footnote{
In the case of no uncertainty, we did not add any extra noise to the system dynamics. However, the vector of variances of the added Gaussian noise for each object was $\vec{\beta} = \{ 0.003,0.006,0.009\}\textbf{1}$ for low, medium and high uncertainty levels respectively. 
} on the velocities (linear and angular) $\vec{v}$ of the robot and dynamic objects at every simulation time step. 
\begin{align}
\label{eq:simulation_noise}
\tilde{\vec{v}} = {\vec{v}} + \vec{\mu} , \hspace{0.5cm} \vec{\mu} \sim \mathcal{N}(0,\vec{\beta})
\end{align}
\noindent where $\mathcal{N}$ is the Gaussian distribution and $\beta$ is the vector of variances.  We give each planner a timeout of 15 minutes, which includes
all planning, re-planning, and execution times. A planner may return long
before this timeout, if the robot manages to grasp the target object in the
execution world. We run and compare the following planners:
\begin{itemize}
\item NR: The naive re-planning algorithm, with $ManyIter=50$, $\nu=0.008$, $K=8$. 
\item OR: The online re-planning algorithm, with
$ManyIter=50$, $FewIter=1$, $n_{min}=2$, $\nu=0.008$, $K=8$, $SD_{thresh}=0.5$.  
\end{itemize}
For all planners, we initialize the control sequences to straight line
trajectories toward the goal. Each initial control sequence includes six
actions, with an average resultant velocity of $0.04 m/s$. Each action is executed for $\Delta_{t}=1 s$.  The
weights and constants used in the cost terms are: $w_g=10000$, $w_{\phi}=1.0$, $w_e=1.0$,
$k=1000$, $w_a=0.1$, $w_d=800$.

\subsection{Simulation Results}

We discuss and compare the performance of OR and NR. 

\begin{figure*}
	\begin{subfigure}[b]{0.245\textwidth}
		\includegraphics[height=1.35in, width=1.75in, angle=0]{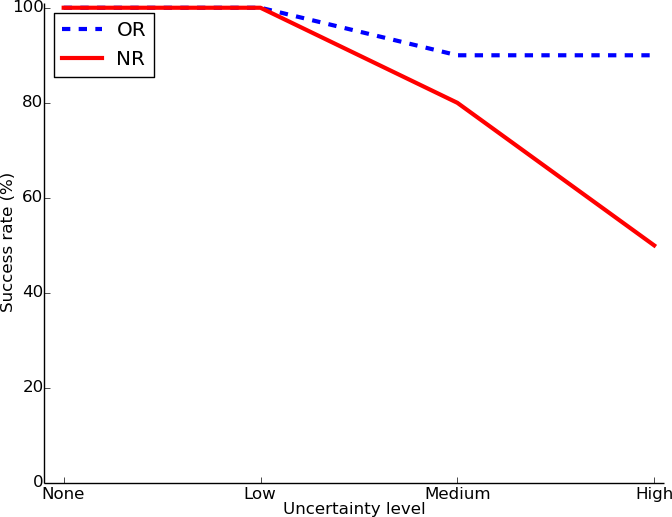}
		\caption{Success rate}
		\label{fig:success_rate}
	\end{subfigure}
	\begin{subfigure}[b]{0.245\textwidth}
		\includegraphics[height=1.35in, width=1.75in, angle=0]{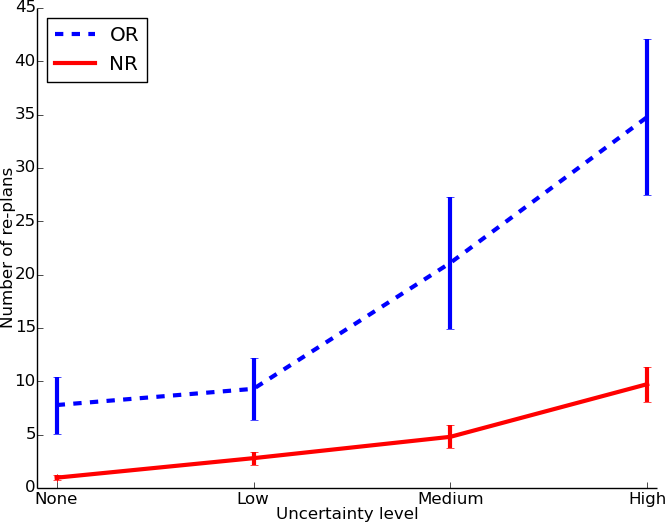}
		\caption{Average number of re-plans}
		\label{fig:num_replans}
	\end{subfigure}
	\begin{subfigure}[b]{0.245\textwidth}
		\includegraphics[height=1.35in, width=1.75in, angle=0]{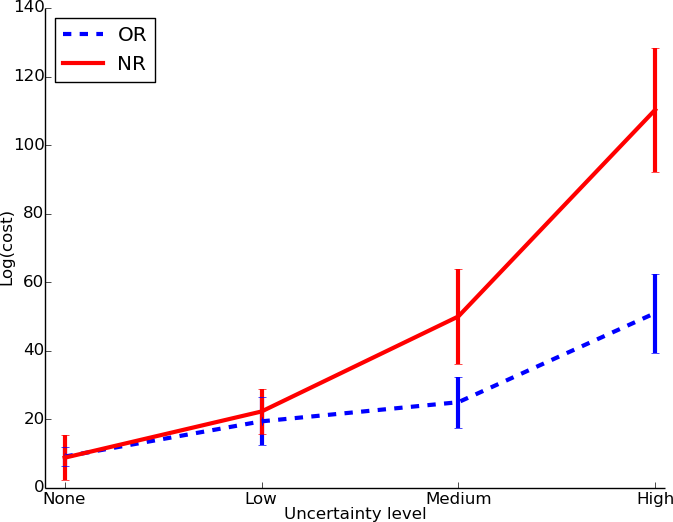}
		\caption{Average execution cost}
		\label{fig:exec_cost}
	\end{subfigure}
	\begin{subfigure}[b]{0.245\textwidth}
		\includegraphics[height=1.35in, width=1.75in, angle=0]{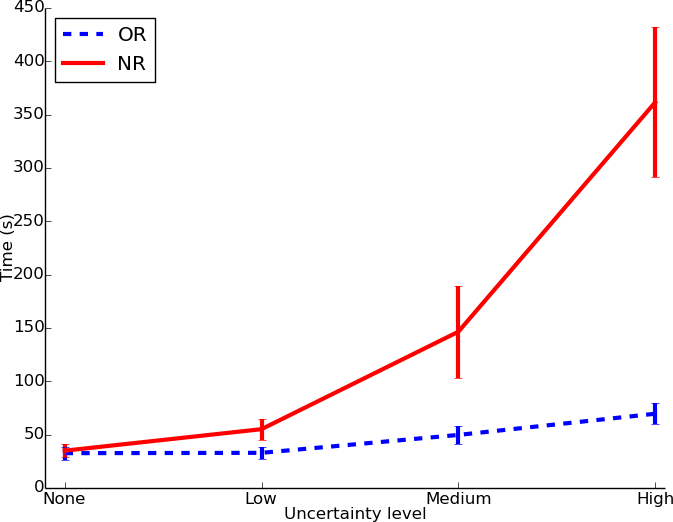}
		\caption{Average elapsed time}
		\label{fig:elapsed_time}
	\end{subfigure}
	\caption{Simulation results for 100 random scenes. In b-d, we plot the average with 95\% confidence interval of the mean}
\end{figure*}

\vspace{3mm}
\noindent\textbf{\textit{OR is more successful than NR.}}  We call an
experiment success, if the execution stopped with the target object inside the
hand pre-grasp region and if no other object is dropped off the table. We show the
success rates over the 100 random scenes under four different uncertainty
levels in Fig.~\ref{fig:success_rate}.  Both OR and NR succeed in all scenes for the
no uncertainty and low uncertainty conditions. However, as the uncertainty
increases, NR shows a dramatic drop to $50\%$ success rate, while OR can
maintain $90\%$.

We show example plans in Fig.~\ref{fig:examplesims}. In the top row, we show
the output of the planner and the state sequence as predicted by the planner in
the Planning World.  In the middle row, we show the NR execution of the same
scene in the Execution World with noise added at the medium uncertainty level. As the hand pushes on a cylinder, it does not move out of the way as the planner predicted. It pushes and topples the target object, resulting in a failure. In the bottom row, we show the OR execution in the same Execution
World. Detecting that the cylinder does not move as predicted, OR re-plans and shifts the gripper to the side, so that the cylinder can be pushed out of
the way.

It is important to note that, the success rates in Fig.~\ref{fig:success_rate}
are not attained after one planning and execution cycle. In other words, the
$100\%$ success rate for NR under low uncertainty does \textbf{not} mean that
open-loop executions of all plans were successful in this case. Instead, it is
more often that the execution of an open-loop plan fails, but leaves the robot
at a close enough point to the target that, the subsequent plans achieve
success.  We present Fig.~\ref{fig:num_replans} to explain this, which shows
the average number of re-plans of each planner under varying uncertainty. As
can be seen, both planners show increasing number of re-plans with increasing
uncertainty. Although, each re-plan is much cheaper for online re-planning compared to naive re-planning. 

\begin{figure*}
	\begin{subfigure}[b]{0.245\textwidth}
		\includegraphics[height=1.35in, width=1.75in, angle=0]{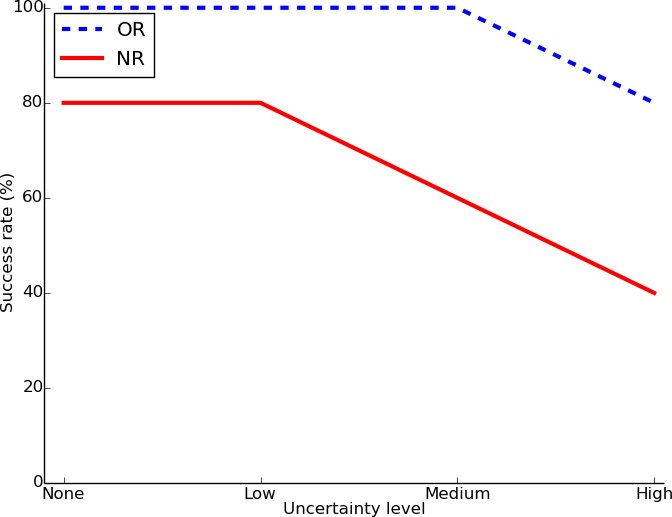}
		\caption{Success rate}
		\label{fig:success_rate_real}
	\end{subfigure}
	\begin{subfigure}[b]{0.245\textwidth}
		\includegraphics[height=1.35in, width=1.75in, angle=0]{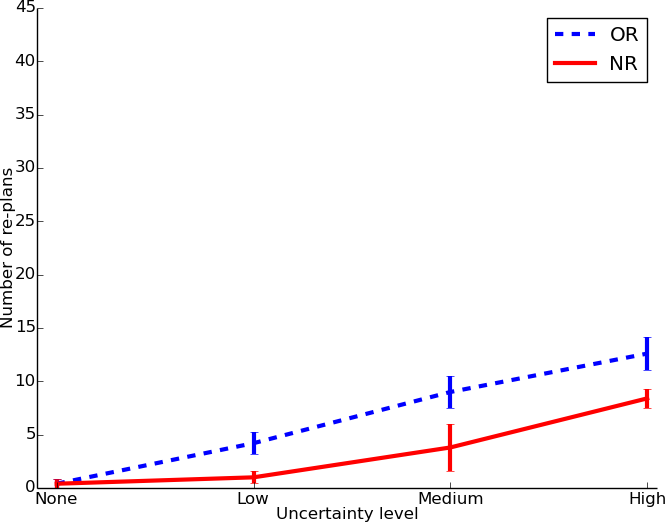}
		\caption{Average number of re-plans}
		\label{fig:num_replans_real}
	\end{subfigure}
	\begin{subfigure}[b]{0.245\textwidth}
		\includegraphics[height=1.35in, width=1.75in, angle=0]{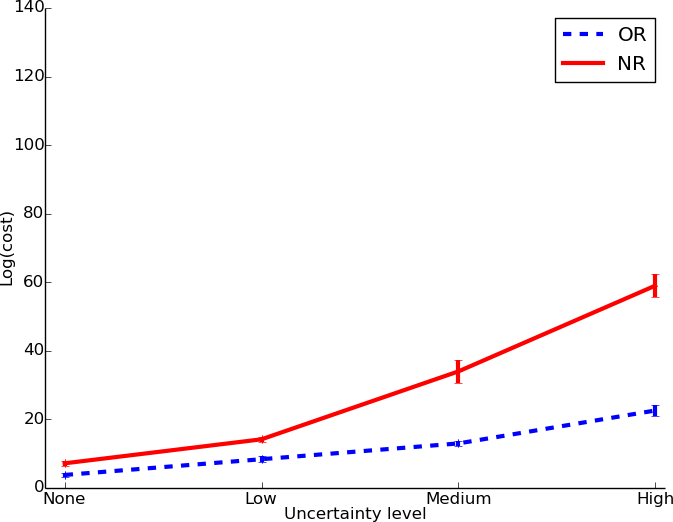}
		\caption{Average execution cost}
		\label{fig:exec_cost_real}
	\end{subfigure}
	\begin{subfigure}[b]{0.245\textwidth}
		\includegraphics[height=1.35in, width=1.75in, angle=0]{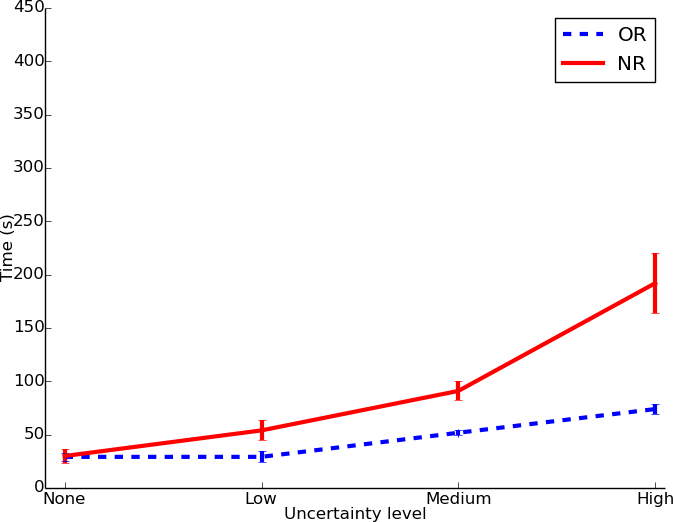}
		\caption{Average elapsed time}
		\label{fig:elapsed_time_real}
	\end{subfigure}
	\caption{Real robot results for 5 random scenes. In (b)-(d), we plot the average with 95\% confidence interval of the mean}
	\label{Fig:real_robot_results}
\end{figure*}

\noindent\textbf{\textit{OR generates lower execution cost than NR.} }
After execution of a planner is completed successfully, we compute the total cost of the executed trajectory. Fig.~\ref{fig:exec_cost} shows the average execution costs
of the planners in log scale versus uncertainty. 
We use only the successful plans for this plot since failure examples run until the arbitrary time limit we have set (15 minutes), and can accumulate arbitrarily large costs.
Again, while OR and NR perform similar at
low uncertainties, the execution cost of NR grows significantly with increasing
uncertainty.

\noindent\textbf{\textit{OR is faster than NR.} }
We record the total re-planning and execution time a planner takes after the
robot makes its first move. Again, we use the successful examples only, since
failure examples run until the pre-set time limit of 15 minutes and therefore do
not give an indication of speed. We plot this total elapsed time in Fig.~\ref{fig:elapsed_time}. Observe that the time NR takes grows rapidly with
uncertainty, while OR is much faster in reaching the goal.

Note that, the above plot does not include the time spent to find the
initial control sequence. We use the PBSTO planner to find this initial
sequence as well for both OR and NR, with a limit of 50 iterations. Averaged over 400 runs
(100 scenes, 4 uncertainty levels), the PBSTO planner needed \textbf{28 seconds} with a standard deviation of 16 to find a plan.

The advantage of the PBSTO planner, however, is that it can also be used
successfully with a small number of iteration limit to quickly adapt plans
under uncertainty. For online re-planning (OR), we ran the PBSTO planner with the iteration limit of 1 for
these quick updates.  During 400 executions, the OR planner performed 7816 such
re-plans. On average, each such update took \textbf{ 0.4 seconds } with a standard
deviation of 0.25 seconds. Therefore, we are able to perform online grasping through clutter in near real time. Moreover, in comparison with works in the literature \cite{randomized_clutter_uncertainty,king2015nonprehensile,kitaev_abbeel,dogar_clutter} about grasping in clutter where the average planning time is in the order of minutes, our approach shows impressive planning and re-planning times.  
\vspace{-0.5mm}
\subsection{Real robot experiments}
In the real robot experiments, we use a Robotiq two finger gripper attached to
a UR5 arm which is then mounted on an omni-directional robot (ridgeback). As shown in Fig.~\ref{fig:fig1}, we fix
the orientation of the arm relative to the table such that is at a specified
height, above the table and is parallel to it. This way the gripper moves
with the omni-directional base yielding a 4 degrees of freedom robot. The
gripper velocities which is the output of our optimization is then transformed
to the omni-directional base through a fixed velocity transform. We place
markers on objects (cylinders and boxes) and sense their full pose (position and
orientation) in the environment using the OptiTrack motion capture system. 

We create $N=5$ execution worlds. We created a mix of difficult (where the target object is behind many closely packed objects) and easy (where the target object is easily accessible) scenes for the experiments. All these scenes can be seen in our video at  \url{https://youtu.be/RcWHXL2vJPc}. Then, we create a planning
world by using estimated values of mass and shape of objects and then get the pose
information from our motion capture system. In addition, we sample the coefficient of friction for the various
objects from a multivariate Gaussian distribution with a mean of
$0.5$ and a variance of $0.01$. 

We are aware that motion capture systems provide a level of object tracking performance which cannot be achieved by using a standard vision system especially in clutter.  Therefore, to see how our online re-planning approach would cope in reality with vision systems, we perform experiments where we artificially insert different levels of pose (x,y positions) uncertainty. We do this by sampling from a Gaussian distribution where the mean is the measured position from our motion capture system. We select a variance of $\{0.005,0.01,0.015\}m$ for low, medium and high uncertainty levels respectively. 

\subsection{Real robot experimental results}
We ran a total of $40$ real robot experiments for $5$ scenes and $4$  uncertainty levels using both naive re-planning and online re-planning. Our results are shown in Fig.~\ref{Fig:real_robot_results}. In general they are similar to the simulation experiments. Moreover, in Fig.~\ref{fig:success_rate_real}, the naive re-planning approach is not always successful even when no artificial uncertainty is added. This is due to the inherent uncertainty in the real world dynamics.

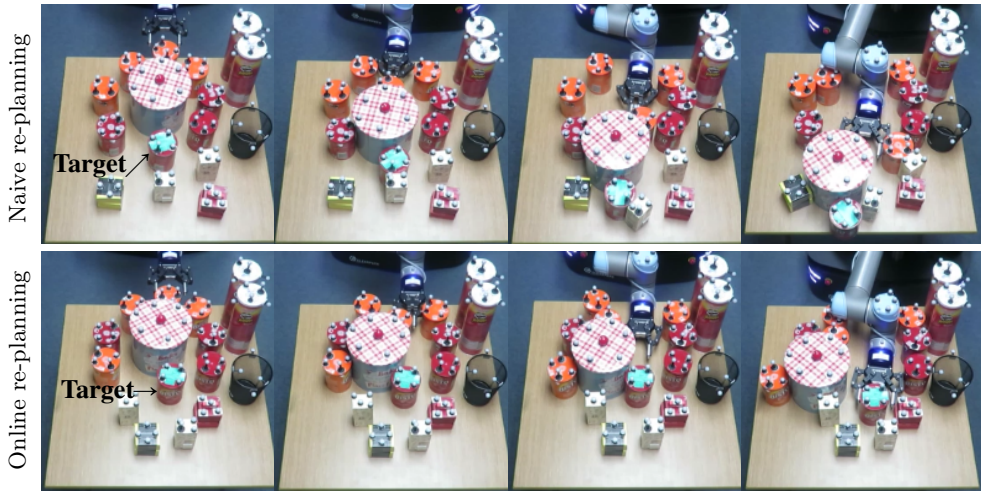
\begin{figure*}[htb!]
\centering 
  \begin{subfigure}[b]{0.245\textwidth}
      \copyrightbox[l]{\includegraphics[height=1.25in,width=1.25in]{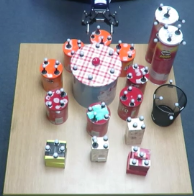}}{Naive re-planning}
      \begin{picture}(0,0)
      \put(-93,27){\textbf{Target}}
      \put(-66,27){\textbf{$\nearrow$}}
      \end{picture}
  \end{subfigure}
  \hspace{-13.0mm}
  \begin{subfigure}[b]{0.245\textwidth}
    \copyrightbox[l]{\includegraphics[height=1.25in,width=1.25in]{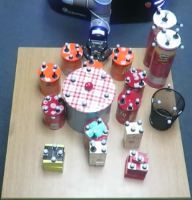}}{}
  \end{subfigure}
   \hspace{-15mm}
  \begin{subfigure}[b]{0.245\textwidth}
    \copyrightbox[l]{\includegraphics[height=1.25in,width=1.25in]{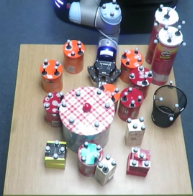}}{}
  \end{subfigure}
   \hspace{-15mm}
  \begin{subfigure}[b]{0.245\textwidth}
    \copyrightbox[l]{\includegraphics[height=1.25in,width=1.25in]{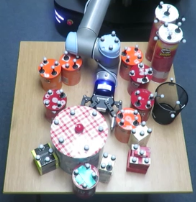}}{}
  \end{subfigure} 
  
	\vspace{-4.5mm}
	
  \begin{subfigure}[b]{0.245\textwidth}
      \copyrightbox[l]{\includegraphics[height=1.25in,width=1.25in]{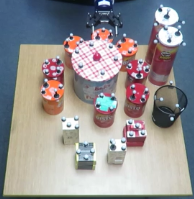}}{Online re-planning}
      \begin{picture}(0,0)
      \put(-90,35){\textbf{Target}}
      \put(-63,35){\textbf{$\rightarrow$}}
      \end{picture}
  \end{subfigure}
  \hspace{-13.0mm}
  \begin{subfigure}[b]{0.245\textwidth}
    \copyrightbox[l]{\includegraphics[height=1.25in,width=1.25in]{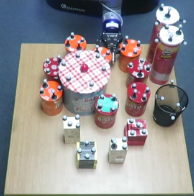}}{}
  \end{subfigure}
     \hspace{-15mm}
  \begin{subfigure}[b]{0.245\textwidth}
    \copyrightbox[l]{\includegraphics[height=1.25in,width=1.25in]{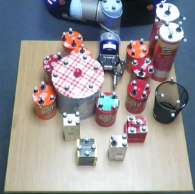}}{}
  \end{subfigure}
     \hspace{-15mm}
  \begin{subfigure}[b]{0.245\textwidth}
    \copyrightbox[l]{\includegraphics[height=1.25in,width=1.25in]{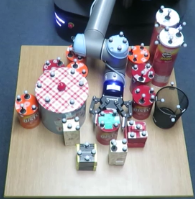}}{}
  \end{subfigure} 
	
\caption{Top row: Naive re-planning (no added uncertainty) fails to grasp the target. Bottom row: Online re-planning succeeds. }\label{fig:examplerob}
\vspace{-3mm}
\end{figure*}

In Fig.~\ref{fig:examplerob}, we show an example scene from our real robot experiments. The naive re-planning approach (top row) was not successful in grasping the target object even under no additional uncertainty. The reason for this is the inherent uncertainty in the real world. More specifically, it is due to the mismatch between the planning environment in simulation and the real world especially in terms of object shape, mass, and friction coefficient. Moreover, the real objects are not fully rigid bodies. Hence predictions of physics in the real world becomes difficult especially for cases where the robot pushes on multiple objects in contact with each other (second snapshot, top row). Therefore, at the end of an open-loop execution in the real world, the robot can put the state of the system in a dead-end (fourth snapshot, top row) from which recovery and task completion becomes extremely difficult. On the other hand, our online re-planning approach shown in the bottom row succeeds in this scene. It is able to track changes between a planned trajectory and the actual state trajectory in the real world. We re-plan if the changes are large and continue this process until the robot successfully grasps the target object. Videos of sample executions can be found at  \url{https://youtu.be/RcWHXL2vJPc}. 

\section{Related Work}
Uncertainty is inevitable during non-prehensile manipulation Yu et al. \cite{million_ways_to_push}. One approach to handling uncertainty is through actions that funnel uncertainty to the goal state(s)
\cite{erdmann_mason,berenson_kuffner,stulp,convergent_planning}.
Uncertainty is also tackled through using sensor feedback during manipulation. Lynch et al. \cite{lynch_pushing} proposed a pushing control system where tactile feedback and object motion predictions are used. Hsiao et al. \cite{hsiao_perez} describes the manipulation problem as a partially observable Markov decision process (POMDP) and formulates methods to efficiently generate robust strategies. More recently, Zhou et al. \cite{zhou_mason}, proposed a probabilistic algorithm that generates sequential actions to iteratively reduce uncertainty until an object's pose is uniquely known. Hogan and Rodriguez \cite{pusher_slider} investigated the pusher-slider system and proposed a method to push a single object using model predictive control and integer programming. Arruda
et al. \cite{mppi_push} proposed the use of a learned pushing model and model predictive control to push a single object to a goal location. 

Clutter is another challenge that we encounter during manipulation.  Stilman et al. \cite{stilman_asfour}, investigated the manipulation planning amongst movable obstacles problem. More recently, Haustein et al. \cite{haustein_asfour} considered the rearrangement planning problem  as a search for dynamic transitions between statically stable states in the joint configuration space. Dogar et al. \cite{dogar_clutter} proposed a framework for push-grasping in clutter, where a grasp approach trajectory is planned by keeping track of movable objects in the environment. Srivastava et al. \cite{sid_abbeel} integrated a symbolic high-level planner with low-level kinematic trajectory optimization to manipulate objects in clutter. Laskey et al. \cite{laskey_goldberg} proposed the use of a hierarchy of supervisors for learning from demonstrations in order to grasp an object in clutter. Ratliff et al.\cite{chomp} and Schulman et al. \cite{seq_conv_opt} present planners that avoid contact with objects in the environment as they generate plans.  Recently, Kitaev et al. \cite{kitaev_abbeel} proposed physics-based trajectory optimization to handle the clutter-grasping problem. They use the iterative LQR method and define an objective function related to the clutter manipulation problem. As mentioned before, we address a similar problem but different from the literature, we take a closed-loop approach. 
\section{Discussion and Future Work} 
\label{sec:conclusion}
To the best of our knowledge, this is the first work that shows how a robot can complete physics-based manipulation in clutter with online planning in real time. 
Our problem set-up includes many simplifications though. Most importantly, we
do not consider static obstacles which may create jamming effects between the
robot and the objects. In future work, we plan to address this problem, by
extending our planner to handle static boundaries and jamming.

\bibliography{bibliography_file}
\bibliographystyle{IEEEtran}

\end{document}